\documentclass[letterpaper, 10 pt, conference]{ieeeconf}  

\IEEEoverridecommandlockouts                              

\usepackage{graphicx}  
\usepackage{booktabs}  
\usepackage{multirow}  
\usepackage{amsmath}
\usepackage{amssymb}
\usepackage{algorithm}
\usepackage{algpseudocode}
\usepackage{cite}
\usepackage{xcolor}

\newcommand{\sysname}{POSE}  
\usepackage[colorlinks=true, linkcolor=blue, citecolor=red, urlcolor=cyan]{hyperref} 

\title{\LARGE \bf
Pose-aware Legged Robot Semantic Exploration with Omnidirectional Perception in Confined Unknown Environments
}

\author{Xiaoyang Zhan\textsuperscript{*}\thanks{\textsuperscript{*}These authors contributed equally to this work.},
Shiyu Chen\textsuperscript{*}, and Kenji Shimada
\thanks{The authors are with the Department of Mechanical Engineering,
Carnegie Mellon University, Pittsburgh, PA 15213 USA.
{\tt\small xzhan2@andrew.cmu.edu}}
}

\begin{document}

\maketitle
\thispagestyle{empty}
\pagestyle{empty}

\begin{abstract}
Semantic exploration in confined environments requires
both environment mapping and detailed observation of target objects.
For ground robots, limited sensors vertical fields of view and
restricted standoff distances can leave upper object surface
unobserved from planar viewpoints. Body tilting can improve coverage,
but additional observations and posture transitions increase mission
time. To balance this trade-off, we present \sysname{}, a pose-aware semantic exploration system
that exploits a legged robot's intrinsic body pitch and roll with omnidirectional Camera-LiDAR perception. The proposed pose-aware viewpoint
sampling module selects body postures from partial object maps
according to expected coverage gain, while aim-aligned execution
reduces unnecessary body reorientation. Further, we introduce an object-centric viewpoint
pruning strategy assisted by a vision-language model
(VLM), which uses persistent observation history and
bird's-eye-view (BEV) maps to reduce redundant
inspection visits. The resulting semantic viewpoints are combined
with geometric exploration viewpoints in a global exploration planner. Simulations show that \sysname{} improves final target-surface coverage by
8--10 percentage points over the planar planning baseline while
reducing exploration time by 17--32\%, and achieves the highest mean
object coverage AUC among the evaluated
baselines. Real-world experiments with a legged robot carrying an 360 degrees omnidirectional Camera-LiDAR suite
in a machine shop further demonstrate the system's applicability.
These results support adaptive body-posture planning
for improving the coverage--efficiency trade-off
in legged robot semantic exploration. We plan to release the code for community benefit in the future.
\end{abstract}

\section{INTRODUCTION}

Autonomous exploration has been extensively studied
for efficient mapping of unknown environments~\cite{
yamauchi1997frontier,bircher2016receding,tare,gbplanner}.
Applications such as industrial inspection and digital
twins further motivate semantic exploration, which
combines environment mapping with target  observation for tasks ranging from object search
to reconstruction~\cite{SEDEM,starsearcher,find_things,
SB2G,SGE}. Ground robots offer a practical platform
for these tasks, particularly in indoor industrial
environments, where endurance and safe operation
are important.

Previous work for ground robot semantic exploration, in general, selects viewpoints in a planar space\cite{SEDEM, travexplorer, SB2G, SGE, vlfm}. However, this could be insufficient in confined space such as industrial facilities. For example, in narrow aisles, limited standoff distances can leave upper regions of tall objects outside the sensors’ vertical field of view (FOV) at all reachable level-body viewpoints. A legged robot could leverage its body pose, including pitch and roll rotation, to extend its perception limit and address this problem.

\IfFileExists{frontpage.pdf}{%
\begin{figure}[t]\centering
\includegraphics[width=\linewidth]{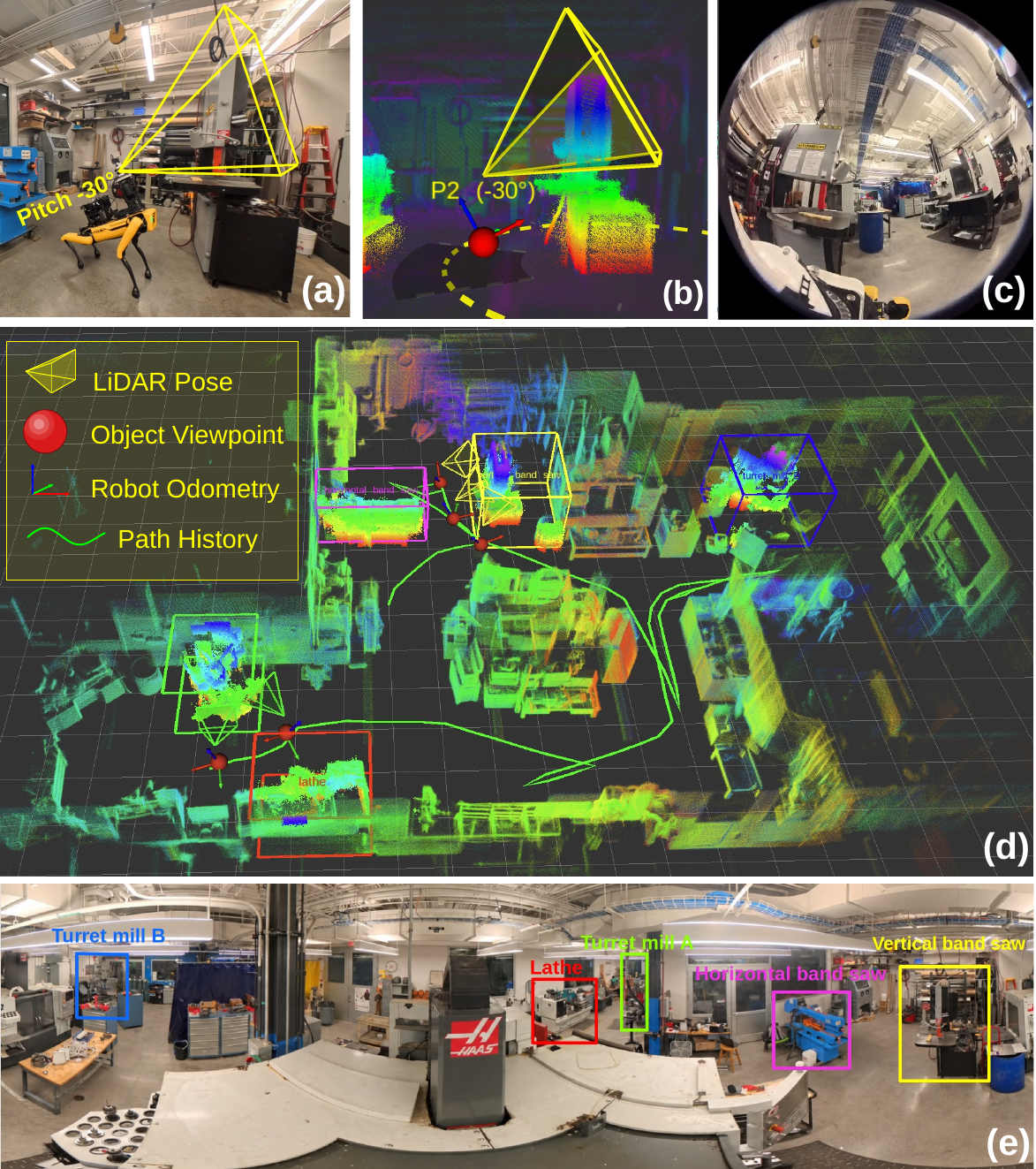}
\caption{\sysname{} inspecting a machine shop with a Boston Dynamics
Spot. (a)~At an object semantic viewpoint, Spot pitches its body up by
$-30^\circ$, raising the LiDAR's FOV onto a machine. (b)~The same viewpoint with its executed pitch in the map. Point cloud of the detected object are colored by height. (c)~The forward camera frame at this posture.
(d)~Overall map: 
accumulated point cloud, detection bounding boxes, object
viewpoints (red spheres), robot odometry at executed
postures, and the traveled path.
(e) A $360^\circ$ photo with every target object boxed in the same color as in (d).}
\label{fig:frontpage}
\end{figure}}{}

Expanding viewpoint planning from planar poses to poses with pitch and roll introduces a trade-off between surface coverage completeness and mission time. Although additional postures can reveal missing surfaces, each execution costs additional time, and a poorly chosen viewpoint with posture costs travel as well without adding coverage. Therefore, a balance between the observation coverage and time efficiency is necessary. This raises two challenges.
First, the selection of observation viewpoints must balance observation richness and efficiency. Omitting posture observations preserves the coverage gap, whereas executing them at every candidate viewpoint introduces redundant travel and posture motion. The planner must decide which viewpoints remain valuable as observations of each object accumulate. This requires considering the object’s observation history when deciding whether another posture observation justifies its cost in time.
Second, posture selection for each viewpoint itself must also account for both the completeness and execution time cost. A newly discovered object is only partially mapped, so its complete model is uncertain. The robot must choose a necessary tilt from this incomplete estimate and refine its decisions as new observations arrive. Also, with omnidirectional perception, different combinations of body heading, pitch, and roll can provide the same observation effect. Therefore, the execution strategy e take advantage of this flexibility to reduce unnecessary reorientation.

To address these challenges, we propose \textbf{\sysname{}}: a \textbf{PO}se-aware legged robot \textbf{S}emantic \textbf{E}xploration system with omnidirectional perception. \sysname{} adopts a hierarchical planner that incorporates pose-aware object semantic viewpoints in a global exploration framework to plan a global tour.
For each target object's semantic observation, \sysname{} proposes a pose-aware viewpoint sampling module with an associated aim-aligned posture execution mechanism. They work together to choose appropriate observation viewpoints with postures and derive an efficient robot body pose for posture execution with less re-orientation.
To reduce unnecessary observation while maintaining observation quality, \sysname{} designs an object-centric VLM-assisted pruning strategy that uses persistent observation context to remove redundant semantic viewpoints.
Our experiments show that across three simulated industrial environments,
\sysname{} improves final target-surface coverage by
8--10 percentage points over the planar exploration baseline
while reducing exploration time by 17--32\%.
It also achieves the highest mean object coverage AUC among
the evaluated baselines, demonstrating its ability
to accumulate greater surface coverage earlier in
the mission. The system is also validated in a university machine shop
(Fig.~\ref{fig:frontpage}). The contributions of this work are:\begin{itemize}
    \item \textbf{A pose-aware semantic viewpoint sampling module} that
extends planar semantic viewpoints with 
body pitch and roll to improve object-surface coverage and
uses aim-aligned posture transitions to reduce reorientation.
    \item \textbf{An object-centric VLM-assisted viewpoint pruning strategy} that takes BEV maps and 
 accumulated observations through persistent VLM sessions
to reduce redundant visits.
    \item \textbf{An integrated semantic exploration and mapping framework} that
jointly optimizes pose-aware semantic inspection and geometric exploration visits with omnidirectional
Camera-LiDAR fusion and real-time semantic mapping.
The system is validated in industrial simulations and
real-world machine-shop experiments.
\end{itemize}

\IfFileExists{methodology.pdf}{%
\begin{figure*}[t]\centering
\includegraphics[width=0.90\textwidth]{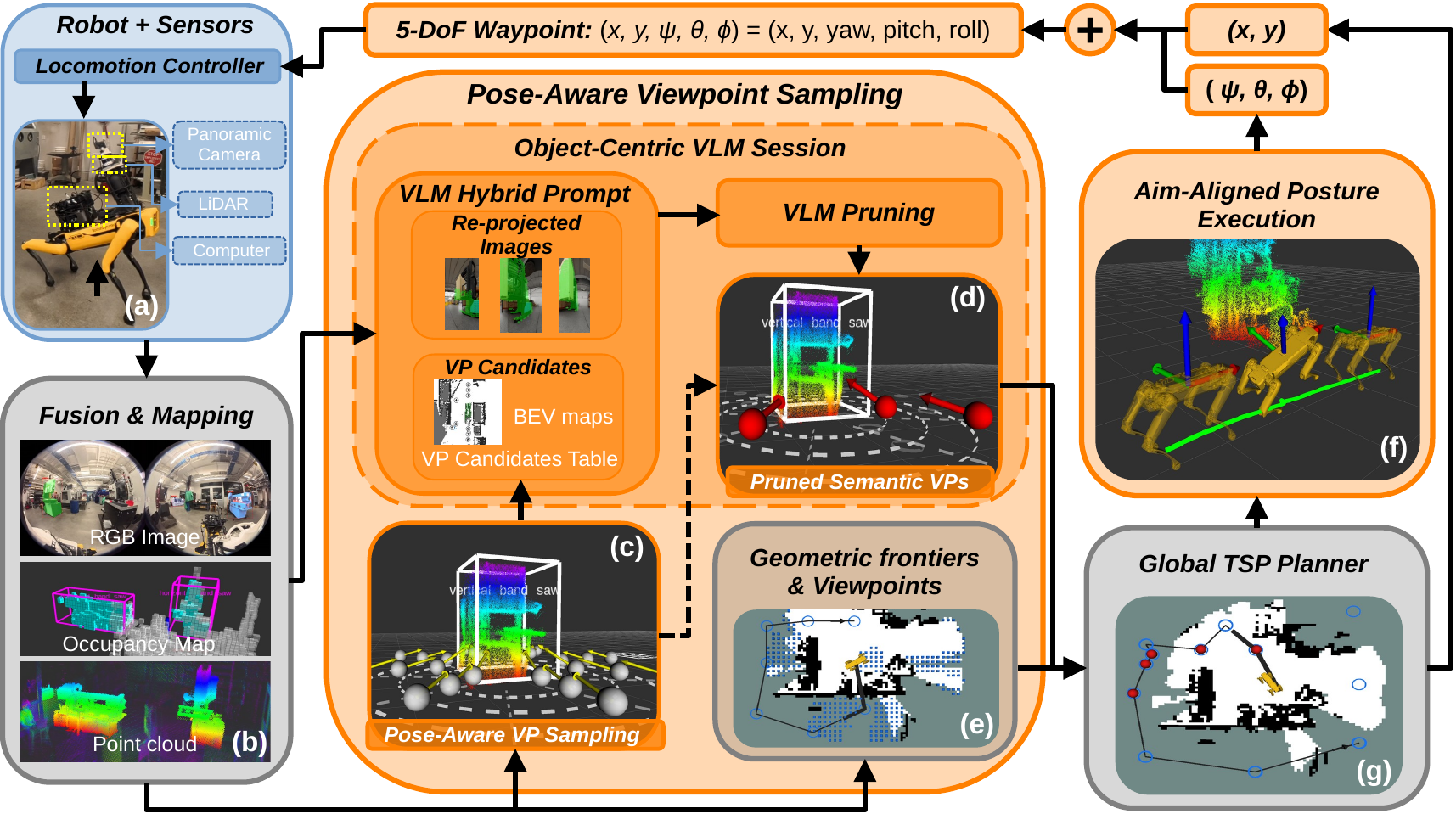}
\caption{Overview of \sysname{}. LiDAR scans and omnidirectional images are
fused into a point cloud and construct, from which an occupancy map with per-object
models is constructed (b). Around each mapped object, semantic viewpoints are
sampled over sectors, standoffs, and postures. A Object-Centric VLM
session is opened upon encountering a new object. Each session consists of 
images, a table and a text prompt. VLM consult and
heartbeat calls return verdicts that prune the viewpoint set. When VLM is not 
available, verdict fallback to geometric default (dashed line). Details are in Sec. \ref{sec:session}.
All viewpoints' visiting sequence is ordered as a TSP (g), and the aim-aligned posture is executed on arrival (f). 
Orange blocks are the contributions of
this paper; gray blocks follow \cite{SEDEM}.}
\label{fig:overview}
\end{figure*}}{}

\section{RELATED WORK}\label{sec:related}
We review exploration planning work for environmental coverage,
semantic exploration,
and active perception methods that exploit additional
sensing degrees of freedom (DoF).

\subsection{Autonomous Exploration}
Autonomous exploration aims to cover unknown space in the shortest time and mainly follows two approaches:
frontier-based methods drive robots toward the boundaries
between free and unknown space~\cite{yamauchi1997frontier},
while sampling-based methods select next-best views
(NBVs) from online graphs~\cite{gbplanner,omniplanner,
bircher2016receding}. Recent planners organize candidate
viewpoints into global coverage tours~\cite{tare,fuel,
dsvp,falcon}. These methods support volumetric exploration
but do not guarantee object-level completeness.

\subsection{Semantic Exploration and Inspection}
Semantic information guides exploration toward different
objectives.
Dang et al.~\cite{semantic_search} prioritize object
observation to detect targets and build coarse semantic
maps during exploration. Luo et al.~\cite{starsearcher}
improve efficiency by jointly planning target search
and environment coverage through global tours.
Hou et al.~\cite{slider} scale search and exploration
to larger environments with lower computational and
memory costs. Ginting et al.~\cite{SB2G} use a semantic
belief behavior graph to locate inspection targets
while reducing uncertainty.

Papatheodorou et al.~\cite{find_things} introduce utility
functions that prioritize camera-based mesh reconstruction
quality during exploration. Dharmadhikari et al.~\cite{swap}
plan for complete mesh reconstruction with image-resolution
and viewing-angle requirements. Zhan et al.~\cite{SEDEM}
coordinate object-observation and geometric-coverage
viewpoints for dense semantic mapping and
efficient exploration. 
Although aerial platforms can adjust viewing altitude,
ground robots in confined spaces may have insufficient
vertical coverage from planar viewpoint changes alone.
However, legged robots can exploit its pitch and roll to offer an additional means of improving target-surface coverage during exploration. 

\subsection{Pose-Aware Viewpoint Planning for Active Perception}
There have been attempts that utilize additional sensing degrees of freedom to obtain task-relevant observations on various platforms.
Hu et al.~\cite{oanbv} sample pitch angles for semantic
observation on a legged robot, without considering
long-horizon exploration. Wang et al.~\cite{wang2020quadruped}
use predefined body-roll sequences for quadruped
inspection. Zhang et al.~\cite{HEATS} combine base--arm
viewpoint configurations for object search and exploration,
with evaluation focused on target-detection completeness
and motion efficiency.
VLM also guides active viewpoint
selection for reconstruction, information gathering,
and question answering~\cite{apvlm,area3d,eyevla}.

 In general, an open challenge still remains: how to exploit the body posture of the legged robot to improve the coverage of semantic objects during exploration of confined unknown environments, while controlling the cost of additional observations. We address this challenge by integrating geometric pitch-and-roll selection, efficient posture execution, and object-level viewpoint pruning with VLM guidance into a unified exploration framework, achieving the balance between surface coverage and exploration efficiency.

\section{Problem Statement}\label{sec:problem}
A quadruped robot explores an unknown, bounded environment
$\mathcal{E}\subset\mathbb{R}^3$ containing an initially
unknown set of targets $\mathcal{O}$ from known semantic
classes. Its sensors comprise a
$360^\circ$ LiDAR with vertical FOV $\pm\varepsilon$
and an omnidirectional camera system.
A viewpoint with posture is
\begin{equation}
v=(x,y,\psi,\theta,\phi),
\label{eq:viewpoint}
\end{equation}
where $(x,y,\psi)$ is the planar pose, and
$\theta\in[-\theta_{\max},\theta_{\max}]$ and
$\phi\in[-\phi_{\max},\phi_{\max}]$ are body pitch and roll.

Observations progressively classify the environment's
voxel set $V$ into free and occupied subsets,
$V_{\mathrm{free}}(t)$ and $V_{\mathrm{occ}}(t)$, at time $t$.
For each target $o$, let $\mathcal{S}_o^\ast$ be its
complete surface voxel set, and $\mathcal{S}_o(t)$ be the reconstructed surface associated with $o$ at time $t$. Define $V_{\mathrm{res}}$ and
$\mathcal{S}_{o,\mathrm{res}}$ as environment voxels
and object-surface voxels, respectively, that cannot
be observed from any reachable, feasible configuration
under the sensor and posture constraints.

The task is to select and execute feasible viewpoints
to explore the environment, discover targets, and
reconstruct their observable surfaces while limiting
travel time. Ideal task completion
at time $T$ requires all the background environment voxels to be observed except  those unobservable, i.e., $V_{\mathrm{free}}(T)\cup V_{\mathrm{occ}}(T)
=V\setminus V_{\mathrm{res}}$, and all the surface of objects to be reconstructed excluding those unobservable, i.e., $\mathcal{S}_o(T)
=\mathcal{S}_o^\ast\setminus\mathcal{S}_{o,\mathrm{res}}$ for $ \forall o\in\mathcal{O}$.

\section{Methodology}\label{sec:method}


\sysname{} comprises semantic mapping and planning pipelines (Fig.~\ref{fig:overview}).
The mapping pipeline fuses omnidirectional camera observations
and LiDAR measurements to build background and object-centric
voxel maps with dense point cloud (Sec.~\ref{sec:base}).
At the global level, the planner iteratively solves an open traveling salesman problem (TSP) over two viewpoint
sets (Sec.~\ref{sec:explore}): exploration viewpoints sampled
around frontiers for volumetric coverage, following~\cite{SEDEM},
and semantic inspection viewpoints sampled around target
objects with geometrically selected body postures for surface
coverage (Sec.~\ref{sec:sampling}).
After viewpoints sampling, the object-centric VLM-assisted pruning
removes redundant inspection candidates using persistent
observation context (Sec.~\ref{sec:session}).
During execution, Aim-Aligned Posture Execution realizes the sampled
scan-band orientation from the arrival heading by re-solving pitch
and roll, avoiding unnecessary re-orientation (Sec.~\ref{sec:exec}).

\subsection{Sensor Fusion and Semantic Mapping}\label{sec:base}
The intrinsics of fisheye cameras are calibrated with the unified spherical camera model, and the extrinsics between
 LiDAR and the two fisheye lenses are calibrated following~\cite{koide2023calib}. 
The LiDAR point cloud and camera RGB images are thus fused given their extrinsics and intrinsics,
yielding a LiDAR depth image registered to
each camera frame. Instance segmentation \cite{yolov8} runs on the fisheye images with a valid detection range of $d_{\max}$. Given the instance mask and LiDAR depth image, the point cloud associated with the target object can be extracted. The details of semantic
mapping are provided in \cite{SEDEM}, which merges detections into
object-centric voxel models with per-sector key-frame statistics. Every
object $o$ exposes its centroid $\mathbf{c}_o$, its box size
$\mathbf{s}_o$, and a dense point cloud registered in the object model.

\subsection{Pose-Aware Semantic Viewpoint Sampling}\label{sec:sampling}
\IfFileExists{gain-score.pdf}{%
\begin{figure}[t]\centering
\includegraphics[width=\linewidth]{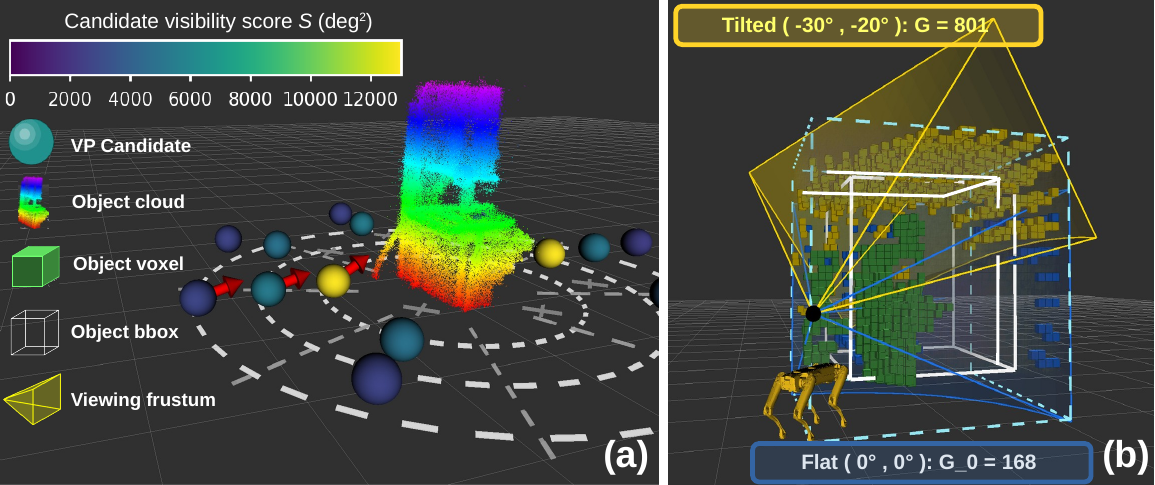}
\caption{Pose-aware semantic viewpoint sampling. (a)~Candidates sampled in sector rings (dashed circles segmented by radial dashed lines)  around a detected object, colored by their visibility score $S$. (b)~Posture gain: the
level band (blue) and the tilted $(\theta^\ast,\phi^\ast)$
(yellow) scan the inflated object box (dashed); the unknown voxels they reach give $G_0$ and $G$, and the tilt is adopted only if $G>G_0$.}
\label{fig:sampling}
\end{figure}}{}

This section describes how semantic inspection viewpoints are
sampled around a mapped object and how each viewpoint receives its
body posture.

\paragraph{Candidate generation}
The space around each object is divided into $N_s$ angular sectors.
In every sector, candidates are sampled on the ray from the object
centroid at a set of standoff distances $\mathcal{D}$, with the base
yaw facing the object, as
Fig.~\ref{fig:overview}(c) illustrates. Candidates with colliding footprints or without valid connecting paths are removed.
A sector is retired once the robot has visited one of its
candidates.

\paragraph{Visibility score}
Each remaining candidate receives a visibility score $S(v)$: the
mapped voxels of the object are ray cast from the candidate, and
$S(v)$ is the panoramic area of the visible ones. Candidates being occluded or
$S(v)$ below a minimum area $S_{\min}$ are discarded. The other candidates 
are kept for the pruning stage of
Sec.~\ref{sec:session}. 

\paragraph{Posture selection}
As shown in Fig.~\ref{fig:overview}(c), each candidate also 
carries a body pitch and roll, chosen on a
discrete grid $\Theta\times\Phi$ that includes the level gaze and
whose extreme values are the body limits $\theta_{\max}$ and
$\phi_{\max}$. The
gain of a posture is the number of unknown voxels inside an inflated
object box $\mathcal{B}_o$ that the tilted sensor band reaches
(Fig.~\ref{fig:sampling}(b)). The object bounding box is expanded by $\delta_b$ to
include potentially unmapped regions. The resulting posture
gain is denoted by $G(v,\theta,\phi)$, which counts unique unknown voxels
within $\mathcal{B}_o$ reached by rays cast through the
sensor FOV at posture $(\theta,\phi)$ with the maximum ray-cast range
$\rho_{\max}$. The selected posture of the candidate is 
\begin{equation}
(\theta^\ast,\phi^\ast)=\arg\max_{(\theta,\phi)\in\Theta\times\Phi}
G(v,\theta,\phi),
\label{eq:posture}
\end{equation}
and it is adopted only if $G(v,\theta^\ast,\phi^\ast)$ exceeds the
posture gain at level-gaze $G_0=G(v,0,0)$. 

The candidates with posture (e.g., white spheres in Fig.~\ref{fig:overview}(c)) 
will be provided to VLM for pruning in Sec. \ref{sec:session}. 


\subsection{Object-Centric VLM-Assisted Viewpoint Pruning}\label{sec:session}
\IfFileExists{session.pdf}{%
\begin{figure}[t]\centering
\includegraphics[width=\linewidth]{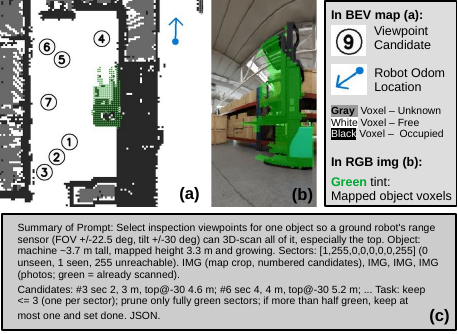}
\caption{One consult input of a VLM session, with example values from a
real run. (a): The top-down map crop around the object with its mapped
voxels and the numbered candidates. (b): The first-person sector images.
(c): The
candidate table (sector, standoff, object height reached at each
tilt angle, visibility score, sector state) and the filled
text prompt.}
\label{fig:session}
\end{figure}}{}
For semantic target observation, visiting each sector's
best viewpoint can introduce redundancy, as accumulated
views may already cover regions visible from unvisited
sectors. Deciding which visits remain useful needs 
interpreting the accumulated visual observations along
with candidate viewing capabilities. \sysname{} thus selects the necessary viewpoints with postures with the assistance of VLM. It maintains a per-object session, which is a persistent asynchronous dialog that
follows the accumulation of observations of the object and returns
keep-only verdicts on its candidate set $\mathcal{V}_o$.
The session issues two kinds of calls: consults, which return
verdicts, and heartbeats, in which a lightweight model checks between
consults whether the current verdict still holds.

\paragraph{Inputs}
Inputs are generated from real-time mapping (Fig. \ref{fig:overview}(b)).
A fixed text template is filled at consult time with the sensor
geometry, the object class and its mapped history, the
observation state of each sector, previous consult records and the candidate table.
Fig.~\ref{fig:session} shows one filled instance. 

Each consult includes up to three images 
(Fig.~\ref{fig:session}(b)). Mapped object voxels are
re-projected onto the sector images as a tint to
indicate observed regions. The candidate
table lists the object height each tilted sensor band
would reach, helping the model identify unscanned
regions and candidates that can observe them. Note that the height is estimated online by given parameters and sensor measurement.
The prompt asks the model to retain candidates that
cover untinted regions and unseen sectors, and to
prune redundant ones. Heartbeats use only the top-down map and the text prompt.

\paragraph{Output}
The response is a JSON with the kept candidates, a confidence score, a
one-sentence reason, and a done flag. An example reply reads:
\begin{quote}\small\texttt{\{"keep": [3], "confidence": 0.76,
"reason": "\ldots a single farther viewpoint that can reach above
4\,m is enough to finish the top coverage.", "done": true\}}
\end{quote}
A guardrail in the prompt counters the tendency of the model to
request more views: once an object is mostly tinted, it is declared
done with at most one further candidate. The posture is not part of
the verdict and remains the geometric choice of Eq.~\ref{eq:posture}.

\paragraph{Receding-horizon session}
The session works in a receding-horizon manner. Each verdict selects semantic viewpoint candidates for one
object and is reassessed as the robot moves.
Algorithm~\ref{alg:session} shows one tick. A consult is called only
in OBSERVING state (Line~11) and the previous consult is passed by 
period $T_c$ (Line~2). The call is asynchronous, so the planner
keeps running on the previous verdict. 
Between consults, each heartbeat asks whether the kept candidates 
still being informative on the updated map. Response other than "ok" push back to OBSERVING state (Line~11). Each verdict expires after a
lifetime $T_{\mathrm{ttl}}$ or when the object centroid drifts beyond
a tolerance. The planner then fallbacks to geometric default candidates (Line~13), which is the best viewpoint in each sector (Sec. \ref{sec:sampling}(c)). If no consult is called after $n$ periods, the session is marked IDLE (Line~12). The prompt in Fig.~\ref{fig:session}(c) carries the previous
verdicts and the consult count, so context persists across ticks.

\begin{algorithm}[t]
\caption{Object-Centric VLM session (one tick)}
\label{alg:session}
\algrenewcommand\algorithmicrequire{\textbf{Input:}}
\begin{algorithmic}[1]
\Require $\eta\in\{\textsc{observing},\textsc{consulting},\textsc{monitoring},$
$\textsc{closed}\}$, last verdict $\nu$
\State cacheObservations() \Comment{crops, map crop, candidate table}
\State \textbf{if} $\eta=\textsc{observing}$ \textbf{and} lastConsultPassed($T_c$) \textbf{then}
\State \quad consultVLM(); $\eta\gets\textsc{consulting}$ \Comment{asynchronous}
\State \textbf{else if} $\eta=\textsc{consulting}$ \textbf{and} replyReady() \textbf{then}
\State \quad $\nu\gets$ parseVerdict(); publish($\nu$); $\eta\gets\textsc{monitoring}$
\State \textbf{else if} $\eta=\textsc{consulting}$ \textbf{and} apiFailed() \textbf{then} 

\State \quad waitSec(120); $\eta\gets\textsc{consulting}$
\State \textbf{else if} $\eta=\textsc{monitoring}$ \textbf{then}
\State \quad status $\gets$ heartbeat()
\State \quad \textbf{if} status $\ne$ ok \textbf{or} newObjectDetected() \textbf{then} 
\State \quad \quad $\eta\gets\textsc{observing}$
\State \textbf{if} objectIdle() \textbf{then} $\eta\gets\textsc{closed}$
\State planner: $\mathcal{V}_o\gets$ keep($\nu$) \textbf{if} \textsc{accept}$(\nu,o)$ \textbf{else} fallback()
\end{algorithmic}
\end{algorithm}

\subsection{Aim-Aligned Posture Execution}\label{sec:exec}
For an ideal z-axis symmetric omnidirectional Camera-LiDAR suite,
different body orientations can produce the same FOV
if the sensor position and scan-band axis remain
unchanged. 
\sysname{} sample a pre-defined candidate yaw angles set, choose the smallest feasible yaw adjustment such that the target scan-band orientation of sensor can be reached. In this way, \sysname{} reduces unnecessary turning angle during posture execution.

With body axes pointing forward, left,
and upward, we use
$R(\psi,\theta,\phi)=R_z(\psi)R_y(\theta)R_x(\phi)$.
The sampled viewpoint
$v=(x,y,\psi_{\mathrm{vp}},\theta^\ast,\phi^\ast)$, whose heading
$\psi_{\mathrm{vp}}$ faces the target, defines the world-frame band axis
\begin{equation}
\mathbf{n}^\ast
=
R(\psi_{\mathrm{vp}},\theta^\ast,\phi^\ast)\mathbf{e}_z,
\label{eq:band_axis}
\end{equation}
where $\mathbf{e}_z=(0,0,1)^\top$. \sysname{} preserves this scan-band axis during execution,
allowing the body roll, pitch, and yaw to change.
For a candidate heading $\psi$, \sysname{} expresses this axis
as $\mathbf{m}(\psi)=R_z(-\psi)\mathbf{n}^\ast$
and recovers:
\begin{equation}
\phi(\psi)=-\arcsin m_y(\psi),\;
\theta(\psi)=\operatorname{atan2}
\bigl(m_x(\psi),m_z(\psi)\bigr).
\label{eq:aim}
\end{equation}
This recovery is exact within
$|\theta|,|\phi|<\pi/2$.

Let $\psi_0$ be the robot's heading upon arrival at
the target viewpoint. The corresponding yaw
adjustment is
$\beta_0=\operatorname{wrap}_{[-\pi,\pi)}
(\psi_{\mathrm{vp}}-\psi_0)$.
\sysname{} searches for the smallest absolute yaw adjustment
that reproduces the sampled scan-band axis within
the pitch and roll limits.
The candidate adjustments are
\begin{equation}
\mathcal{D}_{\psi}
=
\{k\delta_{\psi}:k\in\mathbb{Z},
|k\delta_{\psi}|\le|\beta_0|\}
\cup\{\beta_0\},
\label{eq:yaw_candidates}
\end{equation}
where $\delta_{\psi}$ is the search step.
The selected adjustment is

\begin{equation}
\begin{gathered}
\Delta\psi^\ast
=\arg\min_{\Delta\psi\in\mathcal{D}_{\psi}}
|\Delta\psi|,\\
\text{s.t.}\quad
|\theta(\psi_0+\Delta\psi)|\le\theta_{\max},\\
|\phi(\psi_0+\Delta\psi)|\le\phi_{\max}.
\end{gathered}
\label{eq:turn}
\end{equation}
Candidates are checked in increasing absolute rotation,
starting from zero and searching both directions.
if the search fails, the original target-facing turn $\beta_0$ provides a feasible
fallback
which satisfies the limits by construction.

At the selected heading
$\psi^\ast=\psi_0+\Delta\psi^\ast$,
the robot executes the roll--pitch--yaw command
$(\phi(\psi^\ast),\theta(\psi^\ast),\psi^\ast)$
and holds it for observation.\footnote{
The yaw change could shift the sensor origin by a fraction of the mounting
offset, which is negligible at the standoff distances used, and the
object stays within the omnidirectional camera coverage.}

\subsection{Exploration Planning}\label{sec:explore}
The preceding modules generate posture-aware semantic
inspection viewpoints.
To complement them with volumetric coverage, we sample geometric
exploration candidates in active frontier regions and
connect them to the path-searching graph, following~\cite{SEDEM}.
For each candidate, a path $P=(n_1,\ldots,n_K)$ from the
robot to the candidate is evaluated by
\begin{equation}
r(P)=
\frac{\sum_{j=1}^{K}\gamma^{K-j}g(n_j)}
     {L(P)/u},
\label{eq:exploration_gain}
\end{equation}
where $K$ is the number of path nodes, $n_j$ is the
$j$-th node, and $n_K$ is the candidate viewpoint.
The gain $g(n_j)$ estimates observable unknown space
by ray casting, $\gamma\in(0,1]$ discounts earlier
nodes, and $L(P)/u$ estimates travel time from path
length $L(P)$ and nominal speed $u$.
Intermediate gains are evaluated at spatially separated
locations for efficiency.
Candidates are processed in descending reward order
and retained only if they provide sufficient additional
coverage.

Geometric exploration and retained semantic inspection
viewpoints form a TSP following~\cite{SEDEM}.
The resulting tour is executed through local paths
with periodic replanning in a receding-horizon manner.
The execution state machine permits planning into
unknown space, validates local paths against updated
observations, and triggers re-planning or backward
recovery when execution is blocked.

\begin{table}[t]\centering\small\setlength{\tabcolsep}{2.5pt}
\caption{Summary of Key Parameters}
\label{tab:parameters}
\begin{tabular}{ll}\toprule
Parameter & Value \\
\midrule
$\Theta$ (pitch grid) & $\{-30,-15,0,15,30\}^\circ$ \\
$\Phi$ (roll grid) & $\{-20,0,20\}^\circ$ \\
$N_s$ (number of sectors) & 8 \\
$\mathcal{D}$ (standoff distances) & $\{3,4,5\}$\,m (real: $\{1.2,1.8,2.4\}$\,m) \\
$d_{\max}$ (detection range) & 6.0\,m (real: 3.5\,m) \\
$\rho_{\max}$ (ray-casting range) & 8.0\,m (real: 3.5\,m) \\
$S_{\min}$ & $10^3$\,px$^2$ \\
$\delta_b$ & 0.5\,m \\
Candidates retained per verdict & $\le3$ (real: $\le5$) \\
VLM (consult; heartbeat) & GPT-5.4; GPT-5.4 mini \\
$u$ & 1.2\,m/s (real: 0.8\,m/s) \\
$\omega$ & 0.6\,rad/s (real: 0.5\,rad/s) \\
\bottomrule\end{tabular}\end{table}

\section{Experiments}

\newcommand{\mBase}{SEDEM}
\newcommand{\mGeo}{SEDEM-5DoF}
\newcommand{\mVlm}{VLM-at-Arrival}
\newcommand{\mOurs}{POSE}
\newcommand{\abNoVlm}{w/o VP}
\newcommand{\abNoPost}{w/o PS}
\newcommand{\abNoAim}{w/o AA}
\subsection{Implementation Details}
Simulation experiments use a kinematic Boston Dynamics
Spot model in Isaac Sim with a LiDAR covering
$360^\circ$ horizontally and $\pm22.5^\circ$ vertically,
and two fisheye cameras whose images are combined into
a panorama. Real-world experiments use a Spot with
an Ouster OS-1 LiDAR matching the simulated configuration
and an Insta360 camera, containing of two fisheye lens with a
$210^\circ$ FOV. All computation runs
on a ROG NUC mini PC with an Intel Core Ultra 9 CPU
and an NVIDIA RTX 4070 GPU. For real-world perception,
YOLOv8~\cite{yolov8} provides instance segmentation
and FAST-LIO2~\cite{fastlio} provides real-time dense mapping.
The pruning module accesses VLM inference asynchronously
through an API. Table~\ref{tab:parameters} summarizes
the key planning parameters, shared by simulation
and real-world experiments unless a
real-world value is given in parentheses.

\IfFileExists{all_fig1_cov_vs_time_3x2.pdf}{
\begin{figure}[t]\centering
\IfFileExists{all_fig1_cov_vs_time_3x2.pdf}{\includegraphics[width=\linewidth, height=6cm, keepaspectratio]{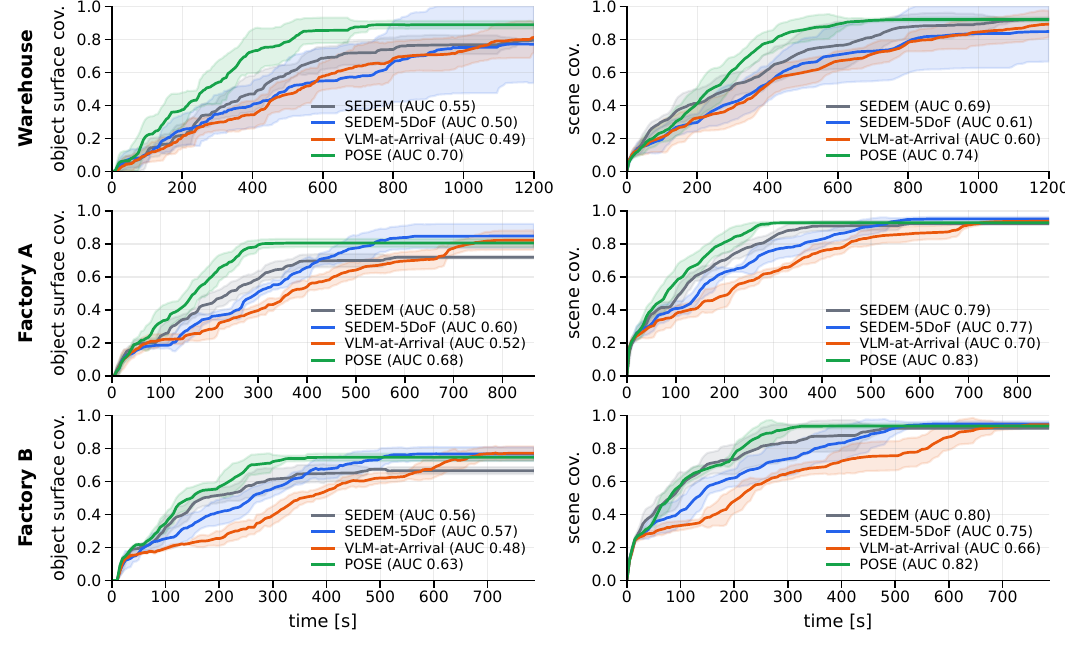}}{}

\caption{Object-surface (left) and scene (right) coverage over
time in Warehouse, Factory A, and Factory B (top to bottom). Curves are
means over valid runs with $\pm 1 \sigma$ bands. The time axis extends
to $T$, the longest run end among all runs of the scene. Methods finishing earlier are held at their final
coverage. Legend entries give the mean per-run object and scene coverage AUC over $[0, T]$, as in Table I.}
\label{fig:cov_vs_time}
\end{figure}}{}

\begin{table*}[t]
\centering
\small
\setlength{\tabcolsep}{5pt}
\caption{Semantic Exploration Benchmark over Three Scenes}
\label{tab:benchmark}
\begin{tabular}{@{}llrrrrrrr@{}}
\toprule
Scene & Method & $AUC_o$ $\uparrow$
& $t_{\mathrm{eff}}$ [s] $\downarrow$
& $L_{\mathrm{eff}}$ [m] $\downarrow$
& $N_{\mathrm{post}}$
& $C_{\mathrm{o}}$ $\uparrow$
& $C_{\mathrm{s}}$ $\uparrow$
& Tokens [k] \\
\midrule
\multirow{4}{*}{Warehouse}
& \mBase
& 0.55\,$\pm$\,0.03
& 812\,$\pm$\,225
& 600\,$\pm$\,137
& 0.0\,$\pm$\,0.0
& 0.79\,$\pm$\,0.01
& \textbf{0.92\,$\pm$\,0.01}
& -- \\
& \mGeo
& 0.50\,$\pm$\,0.18
& 950\,$\pm$\,152
& 563\,$\pm$\,149
& 35.3\,$\pm$\,13.6
& 0.88\,$\pm$\,0.03
& 0.84\,$\pm$\,0.20
& -- \\
& \mVlm
& 0.49\,$\pm$\,0.07
& 1130\,$\pm$\,110
& 566\,$\pm$\,97
& 34.5\,$\pm$\,4.5
& 0.85\,$\pm$\,0.06
& 0.89\,$\pm$\,0.09
& 60.4\,$\pm$\,7.5 \\
& \mOurs
& \textbf{0.70\,$\pm$\,0.05}
& \textbf{614\,$\pm$\,124}
& \textbf{432\,$\pm$\,80}
& 16.3\,$\pm$\,3.0
& \textbf{0.89\,$\pm$\,0.03}
& \textbf{0.92\,$\pm$\,0.02}
& \textbf{28.8\,$\pm$\,4.1} \\
\midrule
\multirow{4}{*}{Factory A}
& \mBase
& 0.58\,$\pm$\,0.01
& 415\,$\pm$\,102
& 261\,$\pm$\,73
& 0.0\,$\pm$\,0.0
& 0.72\,$\pm$\,0.01
& 0.92\,$\pm$\,0.00
& -- \\
& \mGeo
& 0.60\,$\pm$\,0.03
& 545\,$\pm$\,100
& 230\,$\pm$\,42
& 42.7\,$\pm$\,6.7
& \textbf{0.85\,$\pm$\,0.08}
& \textbf{0.95\,$\pm$\,0.02}
& -- \\
& \mVlm
& 0.52\,$\pm$\,0.03
& 750\,$\pm$\,67
& 229\,$\pm$\,32
& 41.8\,$\pm$\,3.4
& 0.82\,$\pm$\,0.06
& 0.93\,$\pm$\,0.02
& 72.4\,$\pm$\,6.4 \\
& \mOurs
& \textbf{0.68\,$\pm$\,0.03}
& \textbf{283\,$\pm$\,46}
& \textbf{174\,$\pm$\,21}
& 12.2\,$\pm$\,2.0
& 0.81\,$\pm$\,0.03
& 0.93\,$\pm$\,0.01
& \textbf{17.7\,$\pm$\,2.7} \\
\midrule
\multirow{4}{*}{Factory B}
& \mBase
& 0.56\,$\pm$\,0.03
& 352\,$\pm$\,95
& 222\,$\pm$\,65
& 0.0\,$\pm$\,0.0
& 0.67\,$\pm$\,0.02
& 0.92\,$\pm$\,0.01
& -- \\
& \mGeo
& 0.57\,$\pm$\,0.03
& 475\,$\pm$\,81
& 199\,$\pm$\,34
& 33.2\,$\pm$\,3.8
& \textbf{0.77\,$\pm$\,0.05}
& \textbf{0.95\,$\pm$\,0.02}
& -- \\
& \mVlm
& 0.48\,$\pm$\,0.02
& 627\,$\pm$\,97
& 226\,$\pm$\,32
& 31.3\,$\pm$\,6.1
& \textbf{0.77\,$\pm$\,0.05}
& 0.94\,$\pm$\,0.01
& 54.4\,$\pm$\,10.2 \\
& \mOurs
& \textbf{0.63\,$\pm$\,0.03}
& \textbf{292\,$\pm$\,52}
& \textbf{173\,$\pm$\,20}
& 9.0\,$\pm$\,0.9
& 0.75\,$\pm$\,0.02
& 0.93\,$\pm$\,0.02
& \textbf{16.5\,$\pm$\,5.2} \\
\bottomrule
\end{tabular}

\end{table*}

\subsection{Simulation Experiments}
We evaluate the planner through benchmark comparisons and ablation
studies in three industrial environments: Warehouse,
Factory A, and Factory B.
The warehouse covers a $39\times57$\,m floor with seven forklifts
about 3.9\,m tall, parked in narrow aisles. Factory A covers
$30\times30$\,m with three forklifts and two stacked crates, all
about 4.0\,m tall. Factory B covers $36\times25$\,m with two machines
and two forklifts of about 4.0\,m. In every scene, the target tops lie above the level sensor band at
the permitted standoffs.
Ground-truth instance segmentation is provided to isolate planning
performance from segmentation errors. All methods share the same
semantic mapping pipeline and termination criteria. A run ends
when scene coverage exceeds $90\%$ and grows by less than
$0.1$ points over a $30\,\mathrm{s}$ window, or when
the time limit of 1200\,s is reached. For finished runs, the effective terminal is the start of the
final $30\,\mathrm{s}$ window, and all metrics use data up to this
cutoff. For timeouts, the cutoff is the time limit.

\IfFileExists{sim-env.pdf}{
\begin{figure*}[t]\centering
\IfFileExists{sim-env.pdf}{\includegraphics[width=0.94\linewidth]{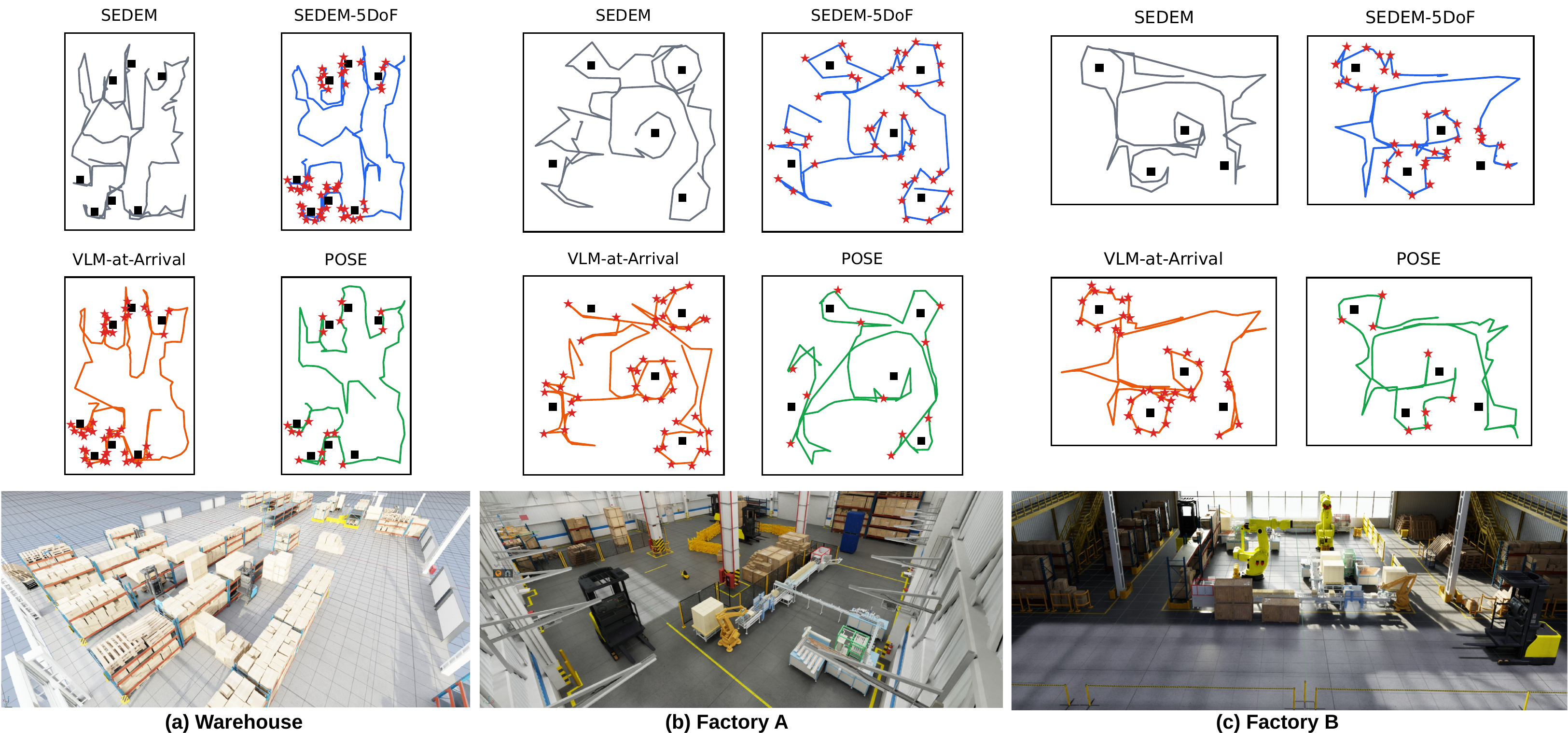}}{}
\caption{Top-down trajectories of the run with median final
object-surface coverage for each method, in Warehouse, Factory A, and Factory B (left to right). Black squares mark target objects; red stars mark executed postures. The bottom row images are screenshots of every scene.}
\label{fig:trajectories}
\end{figure*}}{}

\subsubsection{Benchmark}
We compare \mOurs{} with three baselines. \mBase{}~\cite{SEDEM} is
a recent legged-robot semantic exploration planner using planar
viewpoints. \mGeo{} extends \mBase{} with posture-aware viewpoints:
it samples the same candidates, selects the pose with the highest
unknown-voxel gain without inflating the object bounding box
(Sec.~\ref{sec:sampling}), and faces the object during posture
execution. \mVlm{} applies a naive VLM-based viewpoints posture selection method and follows the same exploration framework without
viewpoint pruning; It captures the
panoramic image, and asks the VLM, the same consult model as
\mOurs{}, for the best viewpoints with postures selections. We report six valid runs per method in each scene, including
timeouts.

We use normalized $AUC_o$ as the primary metric for the trade-off
between object-surface coverage and time. For each run, the $AUC_o$ is
the area under its object-surface coverage curve
(Fig.~\ref{fig:cov_vs_time}, left) on $[0,T]$ divided by $T$, where
the common horizon $T$ is the longest run duration among all methods compared in the same scene.
Shorter runs hold their terminal coverage
until $T$. Higher $AUC_o$ means more coverage achieved earlier.
Fig.~\ref{fig:cov_vs_time} shows object-surface and scene coverage
over time, and Tables~\ref{tab:benchmark} and~\ref{tab:ablation}
report the mean $\pm$ standard deviation over valid runs.
We also report effective exploration time $t_{\mathrm{eff}}$,
effective travel distance $L_{\mathrm{eff}}$, posture executions
$N_{\mathrm{post}}$, object-surface coverage $C_{\mathrm{o}}$, scene
coverage $C_{\mathrm{s}}$, and VLM token usage in thousands. With the
notation of Sec.~\ref{sec:problem},
\begin{equation}
C_{\mathrm{o}}(t)
=
\frac{\sum_{o\in\mathcal{O}}|\mathcal{S}_o(t)|}
     {\sum_{o\in\mathcal{O}}|\mathcal{S}_o^\ast|},
\qquad
C_{\mathrm{s}}(t)
=
\frac{|V_{\mathrm{free}}(t)\cup V_{\mathrm{occ}}(t)|}
     {|V|}.
\label{eq:coverage_metrics}
\end{equation}

Here, $\mathcal{O}$,  $V$ and $\mathcal{S}_o^\ast$ are obtained
from ground truth. The effective time and distance are the elapsed time and the travel distance at the effective
terminal defined above. 

In Table~\ref{tab:benchmark}, \mOurs{} achieves the highest mean
$AUC_o$, the lowest effective exploration time, and the shortest travel
distance in every scene. Compared with \mBase{}, it improves the final
object-surface coverage by 8--10 percentage points while reducing
effective exploration time by 17--32\%, with scene coverage
unchanged. Compared with the two posture-aware baselines, it reaches
comparable object coverage in 35--48\% less time than \mGeo{} and
46--62\% less time than \mVlm{} with 53--73\% fewer postures,
which indicates the effectiveness of our selective posture observation. \mGeo{}
attains 2--4 points higher final coverage in Factory A and B, but at
two to four times the postures and 35--48\% more time. \mOurs{}
also uses 52--76\% fewer VLM tokens than \mVlm{}.
Fig.~\ref{fig:trajectories} shows trajectories and posture
locations.

\subsubsection{Ablation Study}
We evaluate three ablation variants with all other components
unchanged: \abNoVlm{} disables VLM-assisted viewpoint pruning,
\abNoPost{} disables viewpoint posture selection, and restricts all viewpoints to level-body configurations,
and \abNoAim{} disables aim-aligned execution. Thus the robot must face the
object before tilting.
\mOurs{} achieves the highest mean $AUC_o$ in all scenes.
Compared with \abNoVlm{}, pruning reduces posture executions by
55--68\% and effective exploration time by 20--38\% at the best or
comparable object coverage, which confirms the effectiveness of
VLM-assisted pruning. Compared with \abNoPost{}, \mOurs{} improves
final object coverage by 10--12 percentage points and $AUC_o$ by
0.09--0.14, so pose-aware sampling is what raises surface coverage.
Compared with \abNoAim{}, aim-aligned execution saves 3--13\% of
effective time and raises $AUC_o$ by 0.03--0.07 in every
scene; the time differences are within one standard deviation but
consistent in sign.

\begin{table}[t]
\centering
\footnotesize
\setlength{\tabcolsep}{3pt}
\caption{Ablation Study over Three Scenes}
\label{tab:ablation}
\begin{tabular}{@{}llrrrr@{}}
\toprule
Scene & Method & $AUC_o$ $\uparrow$
& $t_{\mathrm{eff}}$ [s] $\downarrow$
& $N_{\mathrm{post}}$
& $C_{\mathrm{o}}$ $\uparrow$ \\
\midrule
\multirow{4}{*}{Warehouse}
& \mOurs
& \textbf{0.70\,$\pm$\,0.05}
& \textbf{614\,$\pm$\,124}
& 16.3\,$\pm$\,3.0
& \textbf{0.89\,$\pm$\,0.03} \\
& \abNoVlm
& 0.61\,$\pm$\,0.02
& 830\,$\pm$\,159
& 36.5\,$\pm$\,5.2
& \textbf{0.89\,$\pm$\,0.03} \\
& \abNoPost
& 0.56\,$\pm$\,0.03
& 684\,$\pm$\,92
& 0.0\,$\pm$\,0.0
& 0.78\,$\pm$\,0.02 \\
& \abNoAim
& 0.63\,$\pm$\,0.05
& 634\,$\pm$\,67
& 15.3\,$\pm$\,2.4
& 0.86\,$\pm$\,0.02 \\
\midrule
\multirow{4}{*}{Factory A}
& \mOurs
& \textbf{0.68\,$\pm$\,0.03}
& 283\,$\pm$\,46
& 12.2\,$\pm$\,2.0
& 0.81\,$\pm$\,0.03 \\
& \abNoVlm
& 0.64\,$\pm$\,0.03
& 455\,$\pm$\,81
& 37.8\,$\pm$\,3.7
& \textbf{0.85\,$\pm$\,0.04} \\
& \abNoPost
& 0.59\,$\pm$\,0.03
& \textbf{278\,$\pm$\,43}
& 0.0\,$\pm$\,0.0
& 0.69\,$\pm$\,0.03 \\
& \abNoAim
& 0.65\,$\pm$\,0.02
& 319\,$\pm$\,73
& 12.8\,$\pm$\,1.5
& 0.79\,$\pm$\,0.02 \\
\midrule
\multirow{4}{*}{Factory B}
& \mOurs
& \textbf{0.63\,$\pm$\,0.03}
& \textbf{292\,$\pm$\,52}
& 9.0\,$\pm$\,0.9
& \textbf{0.75\,$\pm$\,0.02} \\
& \abNoVlm
& 0.59\,$\pm$\,0.05
& 366\,$\pm$\,61
& 26.5\,$\pm$\,7.0
& 0.74\,$\pm$\,0.08 \\
& \abNoPost
& 0.53\,$\pm$\,0.04
& 307\,$\pm$\,93
& 0.0\,$\pm$\,0.0
& 0.65\,$\pm$\,0.01 \\
& \abNoAim
& 0.60\,$\pm$\,0.07
& 300\,$\pm$\,54
& 9.3\,$\pm$\,1.5
& 0.74\,$\pm$\,0.04 \\
\bottomrule
\end{tabular}
\end{table}

\subsection{Real-world Experiments}
We deploy \sysname{} in a machine-shop laboratory (Fig. \ref{fig:frontpage}(e)) measuring about $11.7\times18\times3$ m\, with five machines selected as
semantic targets: an engine lathe, two
turret milling machines, a vertical band saw and a horizontal band
saw, with maximum height around 2.5 m. Multiple runs are conducted, while
 Fig.~\ref{fig:frontpage} presents a representative run. The supplementary video presents an additional
real-world run, illustrating more details about posture execution
and the resulting object-map updates.

During exploration, Spot selects semantic inspection
viewpoints and adjusts its body posture to observe the upper
surface of tall machines. Fig.~\ref{fig:frontpage}(a)
shows an upward-tilted observation, while
Fig.~\ref{fig:frontpage}(b) displays the corresponding
robot pose and the reconstructed object point cloud in RViz.
Fig.~\ref{fig:frontpage}(c) shows the raw image from the
front fisheye camera. The run takes 209.7\,s and a travel
distance of 44.9\,m, with five tilted
postures executed at semantic viewpoints. As shown in Fig.~\ref{fig:frontpage}(d), all five target machines are detected and reconstructed
along with a dense background point cloud, and the red spheres with frustums
show the history of inspection postures.

\section{CONCLUSIONS}
We presented \sysname{}, a semantic exploration system that
selectively exploits the body pitch and roll of a legged robot to
improve object-surface coverage in confined, unknown environments.
It integrates geometric posture selection, object-centric
VLM-assisted viewpoint pruning, and aim-aligned execution into a global exploration planning framework. In three
simulated industrial scenes, \sysname{} improves object-surface
coverage by 8--10 percentage points over the planar baseline while
reducing exploration time by 17--32\%, and achieves the highest mean
coverage $AUC_o$ with 53--73\% fewer postures than the two posture-aware
baselines. Real-world experiments demonstrate its deployability.
These results support selective posture planning for improving the
coverage--efficiency trade-off in legged semantic exploration.


\bibliographystyle{IEEEtran}
\bibliography{IEEEfull.bib}

\end{document}